\documentclass[10pt,twocolumn,letterpaper]{article}

\usepackage{cvpr}              % To produce the CAMERA-READY version
\usepackage{booktabs}
\usepackage{siunitx}
\usepackage{multirow}
\usepackage{graphicx}
\usepackage{pifont}
\usepackage{lscape}

\definecolor{cvprblue}{rgb}{0.21,0.49,0.74}
\usepackage[pagebackref,breaklinks,colorlinks,allcolors=cvprblue]{hyperref}

\def\paperID{4064} % *** Enter the Paper ID here
\def\confName{CVPR}
\def\confYear{2025}

\title{DualTalk: Dual-Speaker Interaction for 3D Talking Head Conversations}

\author{
Ziqiao Peng$^{1}$\quad Yanbo Fan$^{2*\dagger}$\quad Haoyu Wu$^1$\quad Xuan Wang$^2$\quad Hongyan Liu$^3$\quad Jun He$^{1\dagger}$\quad Zhaoxin Fan$^{4\dagger}$ \\
$^1$Renmin University of China\quad $^2$Ant Group\quad  $^3$Tsinghua University \\
$^4$Beijing Advanced Innovation Center for Future Blockchain and Privacy Computing
}

\begin{document}

\twocolumn[{
\maketitle
\begin{center}
    \captionsetup{type=figure}
    \vspace{-6pt}
    \includegraphics[width=1.0\textwidth]{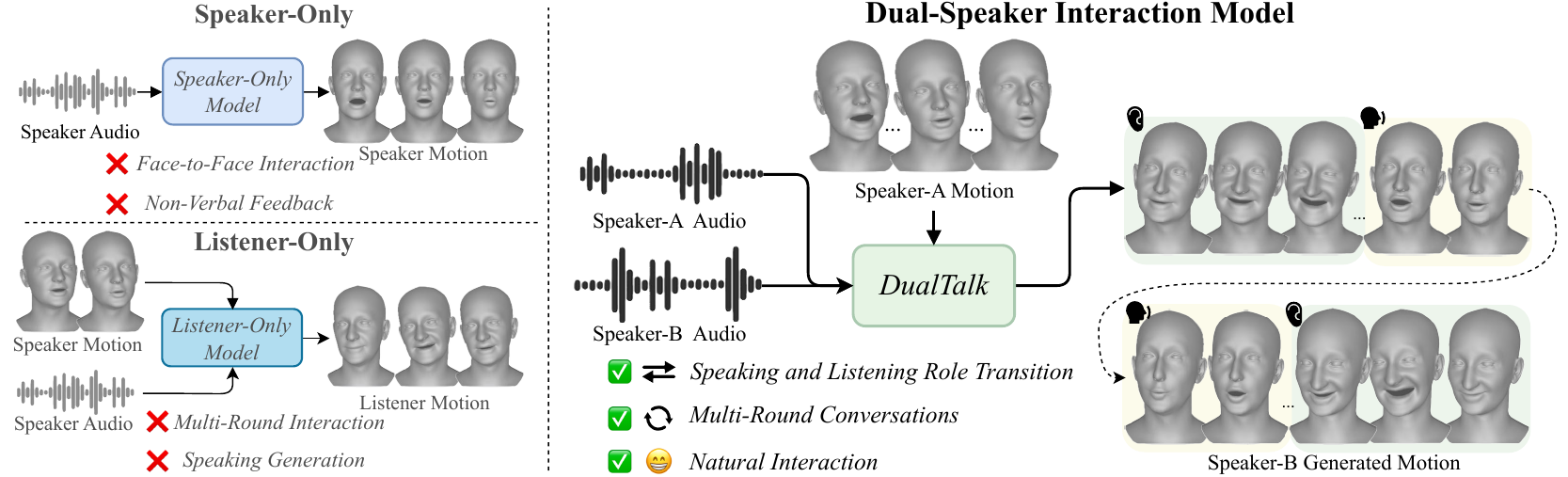}
    \vspace{-1.4em}
    \caption{Comparison of single-role models (Speaker-Only and Listener-Only) with DualTalk. Unlike single-role models, which lack key interaction elements, DualTalk supports speaking and listening role transition, multi-round conversations, and natural interaction.}
    \label{fig:1}

\end{center}
}]

\renewcommand{\thefootnote}{\fnsymbol{footnote}} 
\footnotetext{This work was done during Ziqiao Peng's internship at Ant Group.}
\footnotetext[1]{Project Leader.}
\footnotetext[2]{Corresponding Authors.}

\maketitle
\begin{abstract}

In face-to-face conversations, individuals need to switch between speaking and listening roles seamlessly. Existing 3D talking head generation models focus solely on speaking or listening, neglecting the natural dynamics of interactive conversation, which leads to unnatural interactions and awkward transitions. To address this issue, we propose a new task—multi-round dual-speaker interaction for 3D talking head generation—which requires models to handle and generate both speaking and listening behaviors in continuous conversation. To solve this task, we introduce DualTalk, a novel unified framework that integrates the dynamic behaviors of speakers and listeners to simulate realistic and coherent dialogue interactions. This framework not only synthesizes lifelike talking heads when speaking but also generates continuous and vivid non-verbal feedback when listening, effectively capturing the interplay between the roles. We also create a new dataset featuring 50 hours of multi-round conversations with over 1,000 characters, where participants continuously switch between speaking and listening roles. Extensive experiments demonstrate that our method significantly enhances the naturalness and expressiveness of 3D talking heads in dual-speaker conversations. We recommend watching the supplementary video:\url{https://ziqiaopeng.github.io/dualtalk}

\end{abstract}    
\section{Introduction}
\label{sec:intro}

Interactive conversational agents~\cite{cassell2000human,diederich2022design,liao2023proactive,wu2024vgg}, particularly 3D talking heads~\cite{sung2024laughtalk, li2023one, peng2024synctalk, niswar2009real,zhou2024meta}, are increasingly central to diverse applications, such as customer service, remote work, educational platforms, and entertainment~\cite{zhen2023human,gowda2023pixels,yu2023nofa,pang2023dpe,bai2023high}. The ability of these agents to engage in human-like conversations significantly enhances user experience, offering more intuitive and accessible interactions~\cite{wu2023autogen}. Fluid conversations between participants are crucial, as they make interactions more lifelike and deepen emotional and cognitive engagement.

However, existing 3D talking head methods typically model either the speaker~\cite{zhou2018visemenet, cudeiro2019capture, richard2021meshtalk, fan2022faceformer} or the listener~\cite{ng2022learning, song2023emotional} independently, overlooking the dynamic role-shifting inherent in real-world interactions, where individuals transition seamlessly between speaking and listening. Speaker-only models~\cite{thambiraja2023imitator,xing2023codetalker, peng2023emotalk, peng2023selftalk} generate synchronized lip movements for speaking segments, yet largely neglect the essential listening behaviors that contribute to natural and cohesive interactions. Conversely, listener-only models~\cite{ng2023can,song2023react2023,liu2024customlistener,tran2024dyadic} are often limited to short, reactive expressions, lacking the capacity to capture the ongoing, bidirectional flow of human communication. This gap restricts the authenticity of these conversational simulations.

To bridge this gap, we propose a new task: multi-round dual-speaker interaction for 3D talking head generation. This task emphasizes the limitations of existing speaker-only and listener-only models, which are insufficient to capture the nuanced interplay that shapes the tone, facial expressions, and dynamics of real conversations. For instance, in natural conversations, a speaker’s facial expressions may adjust in response to non-verbal cues from the listener—such as nodding or expressions signaling understanding or confusion. Expanding beyond previous models, our goal is to dynamically simulate both speaking and listening roles, adapting to spoken words as well as non-verbal interactions, thereby enabling more authentic and engaging conversations.

To address this challenge, we introduce DualTalk, a novel unified framework designed to integrate the dynamic behaviors of both speakers and listeners, enabling realistic simulation of multi-round conversational interactions. Unlike previous methods~\cite{ng2022learning,peng2023selftalk, peng2023emotalk} that typically model speaker and listener roles separately, often resulting in static and disjointed interactions, \textbf{DualTalk treats the participant as switching between two states: speaking and listening,} as shown in Fig.~\ref{fig:1}. 
Our approach includes four primary modules to support realistic dual-speaker interactions. The Dual-Speaker Joint Encoder first captures audio and visual signals from each speaker, generating a unified representation. This is followed by the Cross-Modal Temporal Enhancer, which aligns these features over time, preserving the natural flow of conversation. The Dual-Speaker Interaction Module then integrates these features to capture dynamic inter-speaker interplay, allowing for context-aware responses. Finally, the Expressive Synthesis Module fine-tunes the generated expressions, producing nuanced facial animations. This approach not only enhances lip synchronization during speaking segments but also ensures that listener responses are vivid and contextually aligned, capturing the subtle non-verbal cues essential for lifelike interactions.

For training and evaluation, we create a novel dataset for dual-speaker interaction in 3D talking head generation. To the best of our knowledge, this is the first 3D facial mesh dataset crafted for face-to-face, multi-round interactions. This dataset includes dual-channel audio, which allows for isolating each speaker’s voice within multi-speaker environments—a critical feature for analyzing and synthesizing realistic conversations. It comprises approximately 50 hours of conversational data from over 1,000 unique identities, each engaging in multiple rounds of dialogue, with an average of 2.5 rounds per session. Each session is captured with high-quality video, precise audio, and detailed facial expression coefficients. 
With these features, we have constructed a benchmark for evaluating dual-speaker conversations. The DualTalk dataset provides an essential foundation for training models that require rich conversational context and detailed interaction dynamics, enabling the DualTalk model to excel in generating authentic, multi-round conversations.

In summary, the contributions of this paper are as follows:
\begin{itemize}
\item[$\bullet$] We propose a new research task focused on dual-speaker, multi-round interactive conversation, providing a clear framework for modeling continuous dual-speaker dialogues.

\item[$\bullet$] We introduce DualTalk, a unified model designed for multi-round dual-speaker interactions, enabling seamless transitions between speaker and listener roles and enhancing interaction realism.

\item[$\bullet$] We create a novel, large-scale dataset and benchmark specifically designed for dual-speaker interaction, providing an essential foundation for advancing realistic 3D talking head generation in dual-speaker scenarios.

\end{itemize}

\section{Related Work}
\label{sec:related}

\begin{figure*}
\vspace{-1em}
\begin{center}
   \includegraphics[width=1.\linewidth]{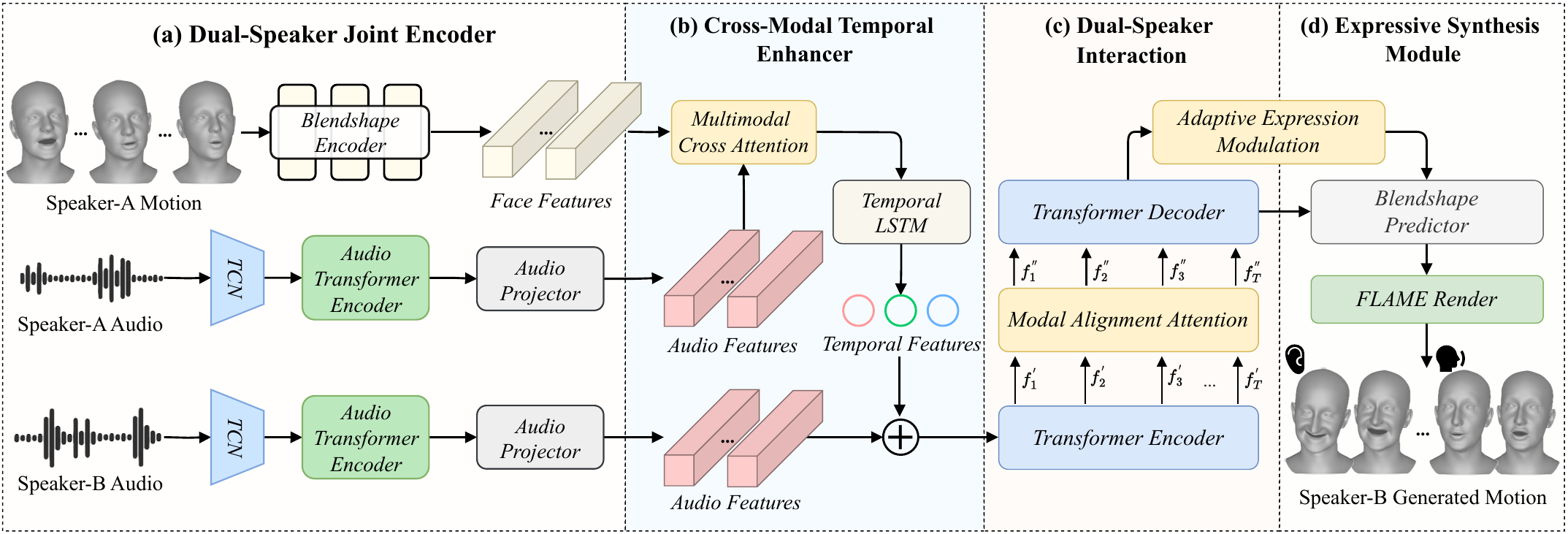}
\end{center}
\vspace{-1em}
   \caption{\textbf{Overview of DualTalk.} DualTalk consists of four components: (a) Dual-Speaker Joint Encoder, (b) Cross-Modal Temporal Enhancer, (c) Dual-Speaker Interaction Module, and (d) Expressive Synthesis Module, enabling the generation of smooth and natural dual-speaker interactions.}
\label{fig:pipe}
\vspace{-0.5em}
\end{figure*}

\subsection{3D Talking Head Generation}

3D talking head generation~\cite{chencafe,zhou2024meta} has become an important research area in computer vision. Early methods primarily focused on generating lip-synchronized facial animations based on audio input. Cao et al.\cite{cao2005expressive} achieved emotional lip synchronization through a constraint-based search within an animation graph structure. This approach required mocap data specific to the animated subject, offline processing, and the combination of various motion segments. Later, Karras et al.\cite{karras2017audio} proposed an end-to-end CNN that learned mapping from waveforms to 3D facial vertices. Recent advancements, such as FaceFormer~\cite{fan2022faceformer}, CodeTalker~\cite{xing2023codetalker}, and SelfTalk~\cite{peng2023selftalk}, introduced geometry-based methods using facial mesh representations to enhance realism in 3D talking heads. UniTalker~\cite{fan2024unitalker} improved generalization by training across multiple datasets and fine-tuning with minimal data, while ScanTalk~\cite{nocentini2024scantalk} enabled 3D face animation with any topology, thus broadening application scenarios. Despite these advancements, most models are still primarily focused on generating individual speech segments, lacking the capacity to support continuous, interactive behaviors required for natural conversation dynamics. 

Our DualTalk model differs from these approaches by jointly modeling both speaker and listener behaviors in dual-speaker scenarios, allowing for seamless transitions between roles.

\subsection{Listener Modeling and Non-Verbal Feedback}

A complementary research area is the modeling of non-verbal listener behaviors~\cite{huang2019toward, zhou2022responsive, ng2023can, ng2024audio, luo2023reactface, liu2023mfr, song2023emotional, song2023react2023,liu2024customlistener, tran2024dyadic}. In human conversations, listeners convey subtle cues through facial expressions, head nods, and eye movements, which play a crucial role in the conversational flow and in making interactions feel more natural. Various studies have modeled listener behaviors with neural networks. For example, Learning2listen~\cite{ng2022learning} generates brief listener reactions, such as head nods and facial expressions, based on the speaker's speech and facial mesh. However, this approach is limited to single-round reactions and does not support continuous, multi-round interactions. Other methods~\cite{liu2024customlistener, song2023emotional, tran2024dyadic} predict facial expressions given a conversational context but capture only brief, isolated reactions, falling short of the fluidity required for extended dialogues.

The work most relevant to our approach is Audio2Photoreal~\cite{ng2024audio}, which generates photorealistic avatars based on conversational audio. However, this model relies solely on audio for modeling and lacks visual feedback from the other participant's expressions. In contrast, our DualTalk method provides a unified framework capable of adjusting based on the counterpart's expressions, enabling seamless role transitions between speaker and listener, thus enhancing the realism and dynamism of interactions across multi-round conversations.

\section{Task Definition}

The primary objective of DualTalk is to generate realistic and dynamic dual-speaker interactions in 3D talking head conversations, enabling natural transitions between speaking and listening roles. Traditional approaches often treat speaker and listener roles separately, failing to capture the fluid dynamics of real-life conversations. Furthermore, without integrated audio-visual understanding, models struggle to adjust a speaker's expressions based on the feedback received from their conversational partner, leading to less natural outcomes. DualTalk aims to address these limitations by simulating responsive, synchronized reactions between two participants.

In this task, the input consists of Speaker-A's audio ($\mathbf{A}_A$) and head motion ($\mathbf{M}_A$), as well as Speaker-B's audio ($\mathbf{A}_B$). Based on these inputs, the model generates Speaker-B's head motion ($\hat{\mathbf{M}}_B$) that synchronizes with the conversational context, reflecting both the verbal and non-verbal cues from Speaker-A. Formally, this process can be defined as a function $f$ mapping the inputs to the desired output:

\begin{equation}
\hat{\mathbf{M}}_B = f(\mathbf{A}_A, \mathbf{M}_A, \mathbf{A}_B),
\end{equation}

where $f$ models the conversational dynamics, allowing Speaker-B's head motion $\hat{\mathbf{M}}_B$ to be conditioned on both Speaker-A's audio and motion, as well as Speaker-B's own audio. This formulation enables DualTalk to generate synchronized and contextually responsive head motions for Speaker-B, effectively capturing the non-verbal feedback and conversational interplay characteristic of natural dual-speaker interactions.

\section{Method}
\subsection{Overview}
In this section, we introduce DualTalk, a unified framework designed to model dual-speaker interactions for 3D talking head generation, as depicted in Fig.~\ref{fig:pipe}. The framework consists of four main components: the Dual-Speaker Joint Encoder, Cross-Modal Temporal Enhancer, Dual-Speaker Interaction Module, and Expressive Synthesis Module. Each component contributes to generating coherent and expressive 3D talking head animations.

\subsection{Dual-Speaker Joint Encoder}

The Dual-Speaker Joint Encoder captures multimodal features from both speakers, integrating audio and blendshape information into a unified feature space. This encoder includes separate Wav2Vec 2.0~\cite{baevski2020wav2vec} audio encoders for each speaker, which process the audio inputs $\mathbf{A}_A$ and $\mathbf{A}_B$ into high-dimensional feature representations. Additionally, the encoder includes a blendshape processing branch that encodes the blendshape parameters, capturing Speaker-A's facial motion  $\mathbf{M}_A$.

Let $\mathbf{A}_A \in \mathbb{R}^{T_A \times F}$ and $\mathbf{A}_B \in \mathbb{R}^{T_B \times F}$ represent the raw audio signals for Speaker-A and Speaker-B, respectively, where $T_A$ and $T_B$ are the sequence lengths and $F$ is the sampling rate. Each audio input is processed through a pre-trained Wav2Vec 2.0 encoder \cite{baevski2020wav2vec}:
\begin{equation}
    \mathbf{H}_A = E_\text{Audio1}(\mathbf{A}_A), \quad \mathbf{H}_B = E_\text{Audio2}(\mathbf{A}_B),
\end{equation}
where $\mathbf{H}_A, \mathbf{H}_B \in \mathbb{R}^{L \times D}$ are the encoded audio features, with $L$ being the length of the encoded feature sequence and $D = 1024$ representing the output embedding dimension from the audio encoder. 

These high-dimensional audio features are then linearly projected into a shared feature space of dimension $d$:
\begin{equation}
    \mathbf{Z}_A = \mathbf{W}_a \mathbf{H}_A, \quad \mathbf{Z}_B = \mathbf{W}_a \mathbf{H}_B,
\end{equation}
where $\mathbf{W}_a \in \mathbb{R}^{d \times D}$ is a learnable projection matrix, and $\mathbf{Z}_A, \mathbf{Z}_B \in \mathbb{R}^{L \times d}$ are the transformed audio features for both speakers, mapped into a lower-dimensional space compatible with the blendshape embeddings.

In parallel, the blendshape encoder processes Speaker-A's facial motion coefficients  $\mathbf{M}_A \in \mathbb{R}^{N \times b}$, where $N$ is the number of frames and $b$ is the number of blendshape coefficients (e.g., $b=56$). The blendshape encoder consists of a two-layer fully connected network with ReLU activations, which transforms the input into an embedding space of dimension $d$:
\begin{equation}
    \mathbf{M}_A' = f_{\text{blend}}(\mathbf{M}_A) = \sigma\left( \mathbf{W}_b^{(2)} \, \sigma\left( \mathbf{W}_b^{(1)} \mathbf{M}_A \right) \right),
\end{equation}
where $\mathbf{W}_b^{(1)} \in \mathbb{R}^{b \times \frac{d}{2}}$ and $\mathbf{W}_b^{(2)} \in \mathbb{R}^{\frac{d}{2} \times d}$ are the weights of the fully connected layers, and $\sigma$ denotes the ReLU activation function. The output $\mathbf{M}_A' \in \mathbb{R}^{N \times d}$ is the blendshape feature embedding, which captures the dynamics of facial movements.

\begin{figure*}
\vspace{-1em}
\begin{center}
   \includegraphics[width=1.\linewidth]{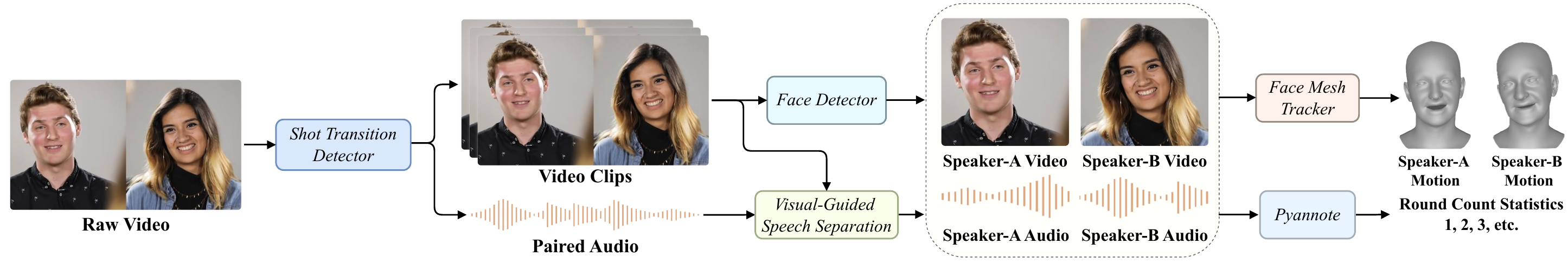}
\end{center}
\vspace{-1em}
   \caption{\textbf{Dataset construction pipeline.} The pipeline takes raw two-speaker videos and paired audio as input. It outputs segmented video clips, isolated audio streams for each speaker, 3D facial mesh data, and speaker round count statistics, providing high-quality, synchronized multimodal data for training.}
\label{fig:dataset_pipe}
\vspace{-0.5em}
\end{figure*}

\subsection{Cross-Modal Temporal Enhancer}

The Cross-Modal Temporal Enhancer module integrates audio and blendshape features, ensuring temporal coherence across frames. This module employs a multimodal cross-attention mechanism to align audio and visual modalities, followed by a bidirectional LSTM~\cite{hochreiter1997long} to capture temporal dependencies. This structure allows for synchronized temporal dynamics, resulting in a coherent multimodal representation across time. The cross-attention mechanism enhances the blendshape features by leveraging audio cues, producing a fused feature $\mathbf{C} \in \mathbb{R}^{L \times d}$.

Specifically, cross-attention is computed as follows:
\begin{equation}
    \mathbf{Q} = \mathbf{Z}_A \mathbf{W}_q, \quad \mathbf{K} = \mathbf{M}_A' \mathbf{W}_k, \quad \mathbf{V} = \mathbf{M}_A' \mathbf{W}_v,
\end{equation}
where $\mathbf{W}_q, \mathbf{W}_k, \mathbf{W}_v \in \mathbb{R}^{d \times d}$ are learnable matrices for query, key, and value vectors. The cross-attention output is computed as:
\begin{equation}
    \mathbf{C} = \text{softmax}\left( \frac{\mathbf{Q} \mathbf{K}^\top}{\sqrt{d}} \right) \mathbf{V}.
\end{equation}
This formulation allows the blendshape features to be modulated by the audio features, aligning the visual representation with the acoustic cues in a contextually-aware manner.

After obtaining the cross-attention output $\mathbf{C}$, the next step is to model temporal dependencies using a bidirectional LSTM. The bidirectional LSTM processes $\mathbf{C}$ in both forward and backward directions, which captures context from both past and future frames:
\begin{equation}
    \mathbf{T} = \text{BiLSTM}(\mathbf{C}).
\end{equation}
$\mathbf{T} \in \mathbb{R}^{L \times 2h}$ represents the temporally enhanced feature, where $h$ is the hidden size of the LSTM. The bidirectional nature of the LSTM allows the model to consider both prior and subsequent context within the temporal sequence, which is crucial for producing a coherent cross-modal output.

Finally, $\mathbf{Z}_A$ and $\mathbf{T}$  are concatenated along the feature dimension to form a combined representation $\mathbf{I} \in \mathbb{R}^{L \times 2d}$:

\begin{equation}
    \mathbf{I} = \text{Concat}(\mathbf{Z}_A, \mathbf{T}),
\end{equation}

where $\text{Concat}(\cdot)$ denotes the concatenation operation along the feature dimension. The resulting $\mathbf{I}$ encodes both the primary speaker's audio information and the cross-modal temporal features of the secondary speaker, effectively capturing the multifaceted aspects of the interaction.

\subsection{Dual-Speaker Interaction Module}

The Dual-Speaker Interaction Module captures and enhances interdependencies between speakers, enabling realistic, context-aware interactions. This module utilizes a Transformer encoder, Modal Alignment Attention, and a Transformer decoder to capture complex dual-speaker dynamics.

The combined features are first processed through a Transformer encoder to capture long-range dependencies and intricate interaction patterns between speakers. This encoder outputs feature representations $\mathbf{f}^{'}_1, \mathbf{f}^{'}_2, \dots, \mathbf{f}^{'}_T$ that encode the dynamics of both speakers across the conversation sequence.

To effectively align these multimodal features, we introduce a Modal Alignment Attention mechanism using an alignment mask, inspired by FaceFormer's biased attention~\cite{fan2022faceformer}. This mechanism adjusts the focus between audio and facial cues, synchronizing responses of both speakers and ensuring contextual alignment in generated interactions. The Modal Alignment Attention (M-A Attention) refines the Transformer encoder outputs to align temporal information from both speakers, enhancing response coherence:
\begin{equation}
\mathbf{f}^{''}_t = \text{M-A Attention}(\mathbf{f}^{'}_t), \quad t = 1, 2, \dots, T
\end{equation}

The refined sequence $\mathbf{f}^{''}_1, \mathbf{f}^{''}_2, \dots, \mathbf{f}^{''}_T$ is then passed through a Transformer decoder, which iteratively processes these features to produce a contextually enriched representation. This representation captures the primary speaker's expressions while dynamically incorporating non-verbal cues from the secondary speaker, facilitating bidirectional interaction. The output of the Transformer decoder, denoted as $\mathbf{D}$, is then passed to the Expressive Synthesis Module.

\subsection{Expressive Synthesis Module}

The Expressive Synthesis Module is the final component responsible for generating the facial animations by predicting blendshape parameters that drive the 3D talking head. 

The Transformer decoder output $\mathbf{D}$ is processed through an adaptive expression modulation mechanism to enhance emotional expressiveness. This step ensures that the final blendshape parameters capture not only lip-sync accuracy but also the emotional tone of the interaction. The adaptive expression modulation is defined as:
\begin{equation}
    \mathbf{D}' = \mathbf{D} + \alpha \cdot \text{Mod}(\mathbf{D}),
\end{equation}
where $\alpha$ is a modulation factor that controls the extent of adjustment, and $\text{Mod}(\mathbf{D})$ is computed as:
\begin{equation}
    \text{Mod}(\mathbf{D}) = \sigma(\mathbf{D} \mathbf{W}_m + \mathbf{b}_m),
\end{equation}
with $\mathbf{W}_m \in \mathbb{R}^{d \times d}$ and $\mathbf{b}_m \in \mathbb{R}^{d}$ as learnable parameters and $\sigma$ as the ReLU activation function. The modulation term dynamically adjusts the expression output based on interaction context, adapting expressions to fit emotional cues.

Finally, the modulated output $\mathbf{D}^{'} \in \mathbb{R}^{L \times d}$ is mapped to the blendshape parameter space through a fully connected layer:
\begin{equation}
    \mathbf{\hat{M}}_B = \mathbf{D}' \mathbf{W}_o + \mathbf{b}_o,
\end{equation}
where $\mathbf{W}_o \in \mathbb{R}^{d \times b}$ and $\mathbf{b}_o \in \mathbb{R}^{b}$ are learnable parameters for the output layer, and $b$ denotes the dimensionality of the blendshape parameters (e.g., $b = 56$). The output $\mathbf{\hat{M}}_B \in \mathbb{R}^{L \times b}$ represents the predicted Speaker-B's face motion coefficients for each frame, which directly controls the 3D facial animation.

\begin{figure}
% \vspace{-1em}
\begin{center}
   \includegraphics[width=1.0\linewidth]{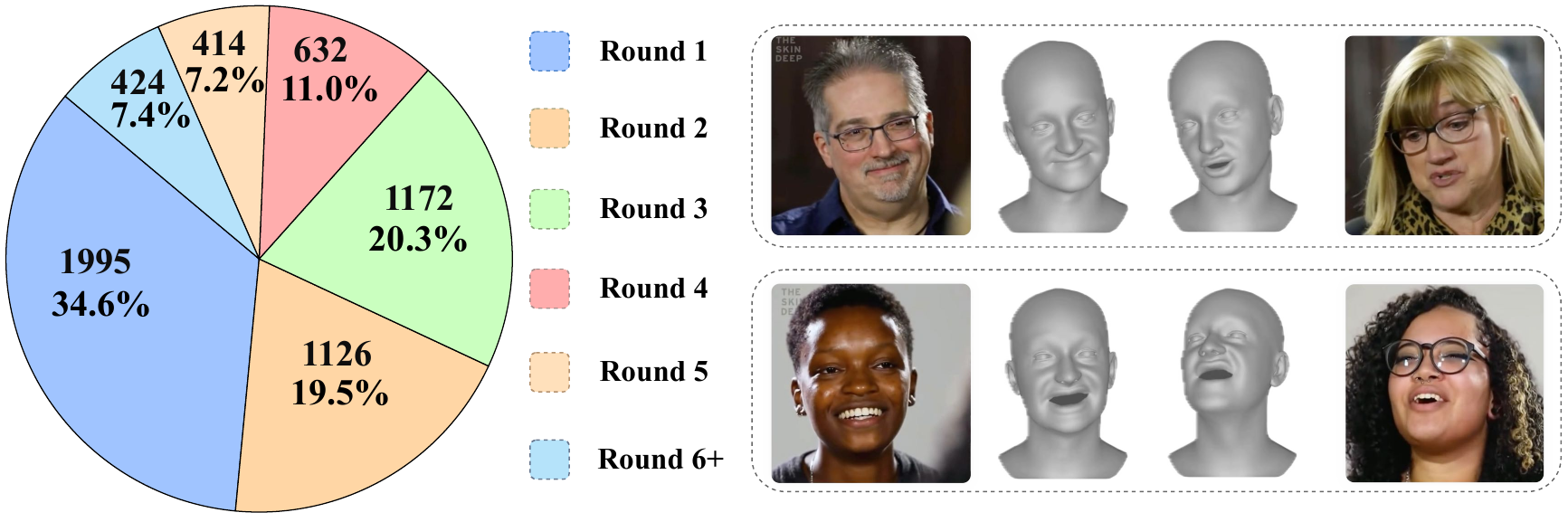}
\end{center}
% \vspace{-1em}
   \caption{\textbf{Distribution} of conversation rounds in DualTalk dataset and example samples.}
\label{fig:dataset_short}
% \vspace{-0.5em}
\end{figure}

\begin{table}[]
\resizebox{\columnwidth}{!}{%

\centering

\begin{tabular}{@{}lcccccc@{}}
\toprule
\textbf{Datasets}     & \textbf{Duration} & \textbf{Identities} & \textbf{Interaction} & \begin{tabular}[c]{@{}c@{}}\textbf{Multi-Round} \\ \textbf{Conversations}\end{tabular} \\ \midrule
VOCASET~\cite{cudeiro2019capture}      & 0.5h         & 12             & \ding{55}        & \ding{55}                     \\
BIWI~\cite{fanelli20103}         & 1.44h           & 14             & \ding{55}        & \ding{55}                               \\
ViCO~\cite{zhou2022responsive}         & 1.6h         & 92             & \ding{51}        & \ding{55}                                 \\
L2L~\cite{ng2022learning}          & 72h          & 6              & \ding{51}        & \ding{55}                                  \\
Lm\_listener~\cite{ng2023can} & 7h           & 4              & \ding{51}        & \ding{55}                            \\
RealTalk~\cite{geng2023affective} & 8h           & -              & \ding{51}        & \ding{55}                              \\
DualTalk     & 50h          & 1000+          & \ding{51}        & \ding{51}                            \\ \bottomrule
\end{tabular}}
\caption{\textbf{Comparison of different 3D talking head datasets.} DualTalk dataset offers over 50 hours of data, 1,000+ identities, interaction, and multi-round conversations.}
\label{tab:dataset_comparison}
\vspace{-10pt}
\end{table}

% \usepackage{booktabs}
\begin{table*}[]
\resizebox{\linewidth}{!}{
\begin{tabular}{@{}lccccccccccccccc@{}}
\toprule
                             & \multicolumn{3}{c}{\textbf{FD} $\downarrow$}                                                                                                                & \multicolumn{3}{c}{\textbf{P-FD} $\downarrow$}                                                                                                               & \multicolumn{3}{c}{\textbf{MSE} $\downarrow$}                                                                                                                 & \multicolumn{3}{c}{\textbf{SID} $\uparrow$}                                      & \multicolumn{3}{c}{\textbf{rPCC} $\downarrow$}                                                                                                           \\ 
\cmidrule(lr){2-4} \cmidrule(lr){5-7} \cmidrule(lr){8-10} \cmidrule(lr){11-13} \cmidrule(lr){14-16} 
\multicolumn{1}{c}{\textbf{Methods}}                 & \textbf{EXP}   & \textbf{JAW} & \textbf{POSE}      & \textbf{EXP}   & \textbf{JAW} & \textbf{POSE}      & \textbf{EXP}   & \textbf{JAW} & \textbf{POSE}    & \textbf{EXP}   & \textbf{JAW}  & \textbf{POSE} & \textbf{EXP}   & \textbf{JAW}   & \textbf{POSE}  \\ 
                                       &                & $\times 10^3$    & $\times 10^2$           &                & $\times 10^3$    & $\times 10^2$           & $\times 10^1$      & $\times 10^3$    & $\times 10^2$          &                &              &               & $\times 10^2$      & $\times 10^1$      & $\times 10^1$        \\ 
\midrule
\multicolumn{1}{l|}{FaceFormer~\cite{fan2022faceformer}}        & 34.90          & 5.40                                                  & \multicolumn{1}{c|}{8.00}                                                  & 34.90          & 5.40                                                   & \multicolumn{1}{c|}{8.00}                                                    & 6.97                                                & 1.80                                                   & \multicolumn{1}{c|}{2.67}                                                    & 0.54          & 0.36          & \multicolumn{1}{c|}{0.50}          & 13.05                                                  & 2.41                                                & 5.27                                                 \\
\multicolumn{1}{l|}{CodeTalker~\cite{xing2023codetalker}}        & 48.57          & 6.89                                                  & \multicolumn{1}{c|}{10.74}                                                 & 48.57          & 6.89                                                   & \multicolumn{1}{c|}{10.74}                                                   & 9.71                                                & 2.29                                                   & \multicolumn{1}{c|}{3.58}                                                    & 0             & 0             & \multicolumn{1}{c|}{0}             & 11.06                                                  & 2.33                                                & 5.11                                                 \\
\multicolumn{1}{l|}{EmoTalk~\cite{peng2023emotalk}}           & 29.86          & 4.33                                                  & \multicolumn{1}{c|}{7.54}                                                  & 30.20          & 4.36                                                   & \multicolumn{1}{c|}{7.58}                                                    & 6.88                                                & 1.76                                                   & \multicolumn{1}{c|}{2.59}                                                    & 2.86          & 1.72          & \multicolumn{1}{c|}{0.98}          & 9.89                                                   & 2.19                                                & 4.94                                                 \\
\multicolumn{1}{l|}{SelfTalk~\cite{peng2023selftalk}}          & 35.77          & 5.49                                                  & \multicolumn{1}{c|}{8.14}                                                  & 35.77          & 5.49                                                   & \multicolumn{1}{c|}{8.14}                                                    & 7.15                                                & 1.83                                                   & \multicolumn{1}{c|}{2.71}                                                    & 2.49          & 1.30          & \multicolumn{1}{c|}{1.28}          & 12.25                                                  & 2.39                                                & 4.70                                                 \\
\multicolumn{1}{l|}{L2L~\cite{ng2022learning}} & 24.61          & 3.69                                                  & \multicolumn{1}{c|}{7.08}                                                  & 24.99          & 3.74                                                   & \multicolumn{1}{c|}{7.13}                                                    & 5.68                                                & 1.48                                                   & \multicolumn{1}{c|}{2.49}                                                    & 2.86          & 1.89          & \multicolumn{1}{c|}{1.19}          & 8.52                                                   & 2.06                                                & 4.11                                                 \\
\multicolumn{1}{l|}{\textbf{DualTalk}}          & \textbf{11.14} & \textbf{1.90}                                         & \multicolumn{1}{c|}{\textbf{3.83}}                                         & \textbf{11.88} & \textbf{1.97}                                          & \multicolumn{1}{c|}{\textbf{3.97}}                                           & \textbf{3.59}                                       & \textbf{1.04}                                          & \multicolumn{1}{c|}{\textbf{1.72}}                                           & \textbf{3.48} & \textbf{2.23} & \multicolumn{1}{c|}{\textbf{1.72}} & \textbf{4.73}                                          & \textbf{1.37}                                       & \textbf{2.38}                                        \\ \midrule
\multicolumn{1}{l|}{FaceFormer~\cite{fan2022faceformer}}        & 35.92          & 5.39                                                  & \multicolumn{1}{c|}{8.60}                                                  & 35.93          & 5.39                                                   & \multicolumn{1}{c|}{8.60}                                                    & 7.18                                                & 1.80                                                   & \multicolumn{1}{c|}{2.87}                                                    & 0.54          & 0.40          & \multicolumn{1}{c|}{0.51}          & 11.71                                                  & 2.16                                                & 5.73                                                 \\
\multicolumn{1}{l|}{CodeTalker~\cite{xing2023codetalker}}        & 50.05          & 6.95                                                  & \multicolumn{1}{c|}{11.66}                                                 & 50.05          & 6.95                                                   & \multicolumn{1}{c|}{11.66}                                                   & 10.01                                               & 2.32                                                   & \multicolumn{1}{c|}{3.88}                                                    & 0             & 0             & \multicolumn{1}{c|}{0}             & 10.24                                                  & 2.18                                                & 5.76                                                 \\
\multicolumn{1}{l|}{EmoTalk~\cite{peng2023emotalk}}           & 34.12          & 4.17                                                  & \multicolumn{1}{c|}{8.59}                                                  & 34.44          & 4.21                                                   & \multicolumn{1}{c|}{8.62}                                                    & 7.73                                                & 1.71                                                   & \multicolumn{1}{c|}{2.94}                                                    & 2.89          & 1.79          & \multicolumn{1}{c|}{0.94}          & 9.44                                                   & 1.96                                                & 5.54                                                 \\
\multicolumn{1}{l|}{SelfTalk~\cite{peng2023selftalk}}          & 36.23          & 5.36                                                  & \multicolumn{1}{c|}{8.89}                                                  & 36.23          & 5.36                                                   & \multicolumn{1}{c|}{8.89}                                                    & 7.24                                                & 1.79                                                   & \multicolumn{1}{c|}{2.96}                                                    & 2.61          & 1.36          & \multicolumn{1}{c|}{1.08}          & 11.26                                                  & 2.13                                                & 5.67                                                 \\
\multicolumn{1}{l|}{L2L~\cite{ng2022learning}} & 30.49          & 3.82                                                  & \multicolumn{1}{c|}{8.56}                                                  & 30.87          & 3.86                                                   & \multicolumn{1}{c|}{8.61}                                                    & 6.87                                                & 1.54                                                   & \multicolumn{1}{c|}{2.98}                                                    & 2.76          & 1.91          & \multicolumn{1}{c|}{1.11}          & 9.02                                                   & 1.94                                                & 4.99                                                 \\
\multicolumn{1}{l|}{\textbf{DualTalk}}          & \textbf{21.71} & \textbf{3.15}                                         & \multicolumn{1}{c|}{\textbf{5.89}}                                         & \textbf{22.56} & \textbf{3.22}                                          & \multicolumn{1}{c|}{\textbf{6.06}}                                           & \textbf{5.97}                                       & \textbf{1.50}                                          & \multicolumn{1}{c|}{\textbf{2.48}}                                           & \textbf{2.98} & \textbf{1.94} & \multicolumn{1}{c|}{\textbf{1.38}}          & \textbf{6.86}                                          & \textbf{1.60}                                       & \textbf{3.28}                                        \\ \bottomrule
\end{tabular}}
\caption{\textbf{Quantitative comparison on DualTalk dataset.} The top half shows results on the DualTalk Test set, and the bottom half shows results on the OOD set. DualTalk outperforms all baselines across most metrics, indicating superior realism, synchronization, and diversity in generated animations.}
\label{tab:metric}
\end{table*}

\subsection{Dataset Construction}

The DualTalk dataset is created to address the limitations of existing datasets that lack support for dual-speaker, multi-round conversations with comprehensive audio-visual synchronization. Current datasets focus on single-speaker scenarios or lack isolated audio streams for each participant, which is essential for training models that simulate both speaking and listening roles. Additionally, most existing datasets do not capture multi-round conversations, which are critical for capturing natural, back-and-forth interactions. To overcome these gaps, we create a dataset specifically designed for dual-speaker interactions, featuring synchronized audio, video, and FLAME-based~\cite{li2017learning} 3D facial data for high-quality training of 3D talking head generation models. The pipeline of dataset construction is shown in Fig.~\ref{fig:dataset_pipe}, and see supplementary materials for details.

The dataset includes 5,858 video clips, amounting to approximately 50 hours of two-person conversation videos, featuring 1,052 unique speakers. Each clip provides clear visual and audio data, allowing for precise audio-visual synchronization. We analyze the dataset's distribution of dialogue rounds (as shown in Fig.\ref{fig:dataset_short}), which reveals a balanced range from single-round to six or more rounds, with an average of 2.5 rounds per clip. This diversity supports training across varying levels of dialogue complexity. Additionally, Tab.\ref{tab:dataset_comparison} compares the DualTalk dataset with other datasets, highlighting its unique advantages. The dataset is divided into train, test, and out-of-distribution (OOD) sets, with 4,935 clips in the train set, 539 clips in the test set, and 384 clips in the OOD set. The OOD set includes speakers not present in the train set, facilitating robust model evaluation.

\section{Experiments}

% \usepackage{booktabs}
\begin{table}[]
\resizebox{\linewidth}{!}{
\begin{tabular}{@{}lccc@{}}
\toprule
\textbf{Methods}         & \begin{tabular}[c]{@{}c@{}}\textbf{LVE}$\downarrow$\\ ($\times 10^{-5}$ mm)\end{tabular}               & \begin{tabular}[c]{@{}c@{}}\textbf{FDD}$\downarrow$\\ ($\times 10^{-7}$ mm)\end{tabular} & \textbf{LRP}$\uparrow$\\ \midrule
VOCA~\cite{cudeiro2019capture}              & 4.9245          & 4.8447          & 72.67\%        \\
MeshTalk~\cite{richard2021meshtalk}          & 4.5441          & 5.2062          & 79.64\%          \\
FaceFormer~\cite{fan2022faceformer}        & 4.1090          & 4.6675          & 88.90\%          \\
CodeTalker~\cite{xing2023codetalker}        & 3.9445          & 4.5422          & 86.30\%          \\
SelfTalk~\cite{peng2023selftalk}          & 3.2238          & 4.0912          & 91.37\%          \\
DiffSpeaker~\cite{ma2024diffspeaker}          & 3.2879          & 4.4031          & 90.81\%          \\
\textbf{DualTalk} & \textbf{2.7944} & \textbf{3.4006} & \textbf{96.69\%} \\ \bottomrule
\end{tabular}}
\caption{\textbf{Quantitative comparison} on VOCASET dataset.}
\label{tab:voca}
\vspace{-10pt}
\end{table}

\subsection{Quantitative Evaluation}

\begin{figure*}
\vspace{-1em}
\begin{center}
   \includegraphics[width=1.\linewidth]{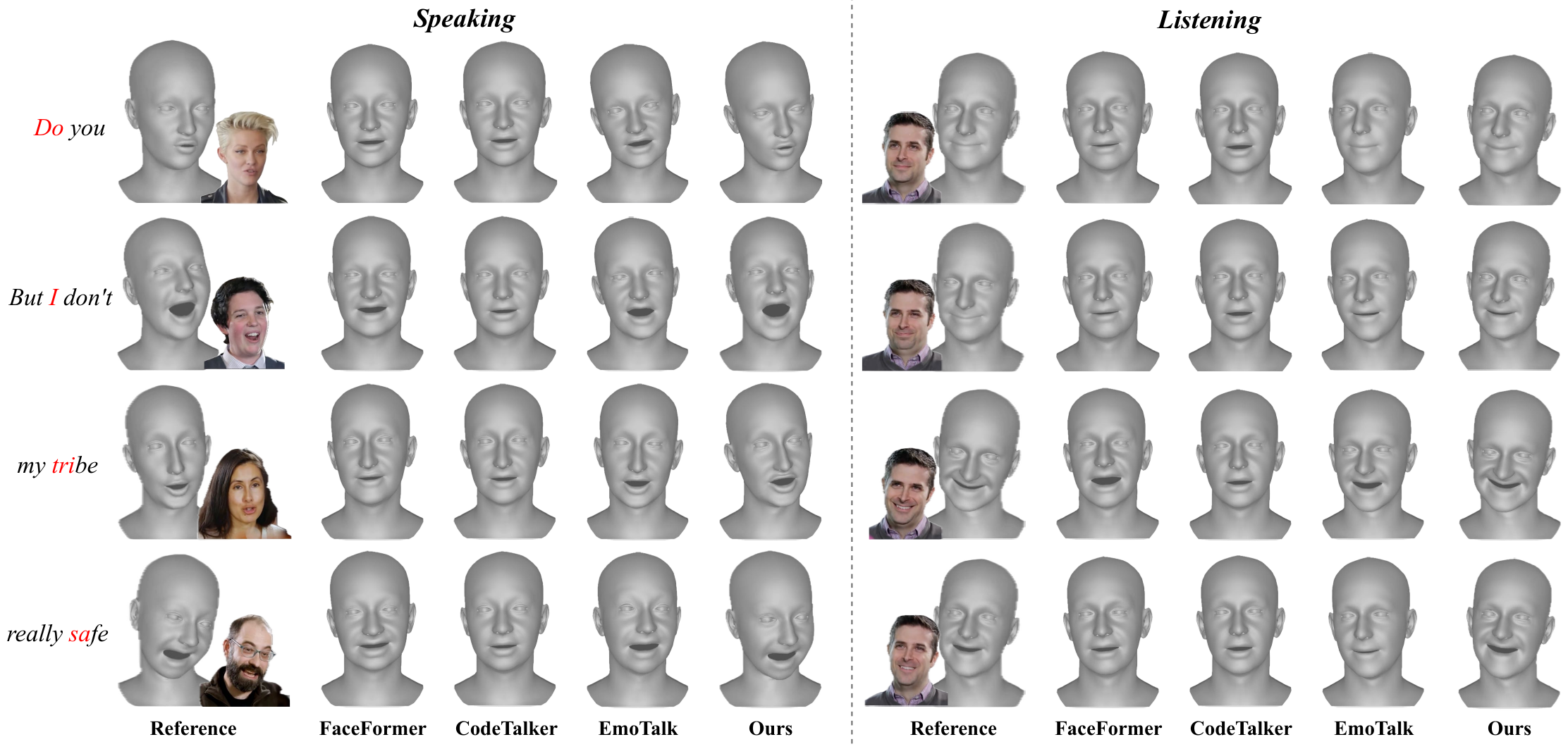}
\end{center}
\vspace{-1em}
   \caption{\textbf{Qualitative comparison of speaking and listening states.} The left side shows facial expressions in the speaking state, with DualTalk achieving more accurate lip movements compared to other methods. The right side shows expressions in the listening state, where DualTalk captures expressive responses like smiling and nodding, outperforming other methods in naturalness and contextual relevance.}
\label{fig:com}
\vspace{-0.5em}
\end{figure*}

We conduct three primary experiments to evaluate the performance of DualTalk, comparing it with baseline models and assessing its effectiveness across different datasets. Detailed experimental settings are provided in the supplementary materials.

\textbf{Baseline Methods on DualTalk Dataset.} In this experiment, we retrain several baseline methods, including FaceFormer~\cite{fan2022faceformer}, CodeTalker~\cite{xing2023codetalker}, EmoTalk~\cite{peng2023emotalk}, SelfTalk~\cite{peng2023selftalk}, and L2L~\cite{ng2022learning}, on the DualTalk dataset. We employ a comprehensive set of evaluation metrics—including Fréchet Distance (FD), Paired Fréchet Distance (P-FD), Mean Squared Error (MSE), SI for Diversity (SID), and Residual Pearson Correlation Coefficient (rPCC)—to assess motion realism, synchronization, diversity, and expression accuracy. As shown in Tab.~\ref{tab:metric}, our method consistently outperforms all baseline models on both the test sequences and out-of-distribution (OOD) sequences of the DualTalk dataset. DualTalk achieves lower errors in FD, P-FD, and MSE, with over a 50$\%$ improvement in expression accuracy compared to the second-best model, L2L~\cite{ng2022learning}. This performance demonstrates that DualTalk generates more accurate facial and head pose movements that better match the dataset’s distribution of head motion. Furthermore, our method improves expression diversity, with a 40$\%$ increase in SID compared to SelfTalk~\cite{peng2023selftalk}, while preserving the movement characteristics of the original dataset. The rPCC metric, which measures motion synchronization between the speaker and listener, shows that our method achieves the best synchronization results. We also test metrics by concatenating outputs from speaker-only and listener-only models, which lead to inferior results. This outcome indicates that DualTalk produces more realistic, synchronized, and diverse motion outputs than other approaches when trained on the same dataset.

% \usepackage{booktabs}
\begin{table}[]
\resizebox{\linewidth}{!}{
\begin{tabular}{@{}lcccccccc@{}}
\toprule
      \multirow{2}[2]{*}{\textbf{Methods}}          & \multicolumn{2}{c}{\textbf{FD} $\downarrow$}         & \multicolumn{2}{c}{\textbf{P-FD} $\downarrow$}     & \multicolumn{2}{c}{\textbf{MSE} $\downarrow$}      \\
               \cmidrule(lr){2-3} \cmidrule(lr){4-5} \cmidrule(lr){6-7}

              & \textbf{exp}   & \textbf{pose} & \textbf{exp}   & \textbf{pose} & \textbf{exp}  & \textbf{pose} \\ \midrule
Random         & 72.88          & 0.12          & 75.82          & 0.12          & 2.05          & 0.03            \\
Nearest Audio  & 65.77          & 0.10           & 68.84          & 0.10           & 1.77          & 0.03              \\
Nearest Motion & 42.41          & 0.06          & 45.33          & 0.06          & 1.27          & 0.02           \\
L2L~\cite{ng2022learning}            & 33.93          & 0.06          & 35.88          & 0.06          & 0.93          & \textbf{0.01}   \\
RLHG~\cite{zhou2022responsive}           & 39.02          & 0.07          & 40.18          & 0.07          & 0.86          & \textbf{0.01}      \\
DIM~\cite{tran2024dyadic}            & 23.88          & 0.06          & 24.39          & 0.06          & 0.70           & \textbf{0.01}      \\
\textbf{DualTalk}       & \textbf{22.27} & \textbf{0.05} & \textbf{23.81} & \textbf{0.05} & \textbf{0.58} & \textbf{0.01}    \\

\bottomrule
\end{tabular}}
\caption{\textbf{Quantitative comparison} on ViCo dataset.}
\label{tab:vico}
\vspace{-10pt}
\end{table}

\textbf{DualTalk on Speaker-Only VOCASET.} To further evaluate DualTalk’s capability in generating high-quality facial animations, we train and test our model on the speaker-only VOCASET. Tab.~\ref{tab:voca} presents results in terms of Lip Vertex Error (LVE), Facial Dynamics Deviation (FDD), and Lip Readability Percentage (LRP). Compared to other audio-driven models, including VOCA, MeshTalk, FaceFormer, and CodeTalker, DualTalk demonstrates significant improvements, achieving the lowest LVE and FDD and the highest LRP score. In particular, we surpass SelfTalk by 5$\%$ in LRP, indicating that our method has superior lip movement accuracy. These results highlight DualTalk’s effectiveness in accurately capturing and replicating audio-driven facial dynamics.

% \usepackage{booktabs}
% \usepackage{graphicx}
\begin{table}[]
\centering
\resizebox{\columnwidth}{!}{%
\begin{tabular}{@{}lcccc@{}}
\toprule
\textbf{Methods} & \begin{tabular}[c]{@{}c@{}}\textbf{Lip Sync} \\ \textbf{Accuracy}\end{tabular} & \begin{tabular}[c]{@{}c@{}}\textbf{Pose} \\ \textbf{Naturalness}\end{tabular} & \begin{tabular}[c]{@{}c@{}}\textbf{Expression} \\ \textbf{Richness}\end{tabular} & \begin{tabular}[c]{@{}c@{}}\textbf{Visual} \\ \textbf{Quality}\end{tabular} \\ \midrule
FaceFormer~\cite{fan2022faceformer} & 2.615 & 2.502 & 2.460 & 2.235 \\
CodeTalker~\cite{xing2023codetalker} & 1.854 & 2.011 & 1.972 & 1.830 \\
EmoTalk~\cite{peng2023emotalk} & 3.250 & 3.471 & 3.331 & 3.269 \\
L2L~\cite{ng2022learning} & 3.872  & 4.067  & 3.814 & 3.750 \\
\textbf{DualTalk} & \textbf{4.164} & \textbf{4.276} & \textbf{4.253} & \textbf{4.088} \\ \bottomrule
\end{tabular}%
}
\caption{\textbf{User Study.} Rating is on a scale of 1-5; the higher the better.}
% \vspace{-10pt}
\label{tab:user}
\vspace{-10pt}
\end{table}
% \vspace{-10pt}

\begin{table*}[]
\resizebox{\linewidth}{!}{
\begin{tabular}{@{}lccccccccccccccc@{}}
\toprule
                             & \multicolumn{3}{c}{\textbf{FD} $\downarrow$}       & \multicolumn{3}{c}{\textbf{P-FD} $\downarrow$}                                                                                                          & \multicolumn{3}{c}{\textbf{MSE} $\downarrow$}                                                                                                                 & \multicolumn{3}{c}{\textbf{SID} $\uparrow$}                                      \\
\cmidrule(lr){2-4} \cmidrule(lr){5-7} \cmidrule(lr){8-10} \cmidrule(lr){11-13}
\multicolumn{1}{c}{\textbf{Ablation Study}}                 & \textbf{EXP}   & \textbf{JAW} & \textbf{POSE}      & \textbf{EXP}   & \textbf{JAW} & \textbf{POSE}    & \textbf{EXP}   & \textbf{JAW}  & \textbf{POSE}  & \textbf{EXP}   & \textbf{JAW}  & \textbf{POSE} \\
                                       &                & $\times 10^3$    & $\times 10^2$        &                & $\times 10^3$    & $\times 10^2$        & $\times 10^1$      & $\times 10^3$    & $\times 10^2$          &                &              &               \\
\midrule
\multicolumn{1}{l|}{\textbf{DualTalk}}        & \textbf{11.14} & \textbf{1.90}                                         & \multicolumn{1}{c|}{\textbf{3.83}}      & \textbf{11.88} & \textbf{1.97}                                         & \multicolumn{1}{c|}{\textbf{3.97}}                                     & \textbf{3.59} & \textbf{1.04}                                          & \multicolumn{1}{c|}{\textbf{1.72}}                                           & \textbf{3.48} & \textbf{2.23} & \multicolumn{1}{c}{\textbf{1.72}}                                           \\
\multicolumn{1}{l|}{w/o Speaker-A’s Speech}        & 23.27          & 4.01                                                  & \multicolumn{1}{c|}{5.48}       & 23.74 & 4.09                                         & \multicolumn{1}{c|}{5.51}                                               & 4.82          & 1.42                                                   & \multicolumn{1}{c|}{1.85}                                                    & 1.68          & 1.23          & \multicolumn{1}{c}{1.13}          \\
\multicolumn{1}{l|}{w/o Speaker-A’s Expression}           & 28.43          & 4.72                                                  & \multicolumn{1}{c|}{5.91}     & 29.10 & 4.79                                         & \multicolumn{1}{c|}{6.05}                                                 & 5.68          & 1.57                                                   & \multicolumn{1}{c|}{1.97}                                                    & 1.47          & 1.05          & \multicolumn{1}{c}{0.97}          \\
\multicolumn{1}{l|}{replace Audio Feature Extractor with MFCC}          & 27.42          & 3.96                                                  & \multicolumn{1}{c|}{7.25}    & 27.95 & 4.02                                         & \multicolumn{1}{c|}{7.31}                                                  & 6.02          & 1.55                                                   & \multicolumn{1}{c|}{2.51}                                                    & 2.71          & 1.66          & \multicolumn{1}{c}{1.06}          \\
\multicolumn{1}{l|}{w/o Cross-Modal Temporal Enhancer}          & 16.31          & 2.91                                                  & \multicolumn{1}{c|}{4.92}      & 16.66 & 2.95                                        & \multicolumn{1}{c|}{4.97}                                                & 4.00          & 1.27                                                   & \multicolumn{1}{c|}{1.79}                                                    & 3.22          & 2.00          & \multicolumn{1}{c}{1.40}          \\
\multicolumn{1}{l|}{w/o Dual-Speaker Interaction Module}        & 16.70          & 2.99                                                  & \multicolumn{1}{c|}{4.78}      & 17.27 & 3.05                                        & \multicolumn{1}{c|}{4.87}                                                & 4.27          & 1.32                                                   & \multicolumn{1}{c|}{1.84}                                                    & 3.03          & 2.05          & \multicolumn{1}{c}{1.47}          \\
\multicolumn{1}{l|}{w/o Adaptive Expression Modulation}          & 13.28          & 2.46                                                  & \multicolumn{1}{c|}{4.53}     & 13.81 & 2.51                                         & \multicolumn{1}{c|}{4.63}                                                 & 3.63          & 1.12                                                   & \multicolumn{1}{c|}{1.80}                                                    & 3.35          & 2.13          & \multicolumn{1}{c}{1.55}          \\
\bottomrule
\end{tabular}}
\vspace{-3pt}
\caption{\textbf{Ablation study for our components.} We show the FD, P-FD, MSE, and SID in different cases.}
\label{tab:ablation}
\vspace{-10pt}

\end{table*}

\textbf{DualTalk on Listener-Only ViCo Dataset.} We evaluate DualTalk on the listener-only ViCo dataset to examine its ability to model listener responses accurately. As shown in Tab.~\ref{tab:vico}, DualTalk outperforms methods such as L2L~\cite{ng2022learning}, RLHG~\cite{zhou2022responsive}, and DIM~\cite{tran2024dyadic}, achieving the lowest FD and P-FD scores, as well as the lowest MSE and highest SID values for listener responses. This performance indicates that DualTalk effectively captures listener-specific head and facial motions, surpassing baseline methods in generating diverse and responsive listener animations.

\textbf{Performance Efficiency.} In addition to accuracy, we evaluate the runtime efficiency of DualTalk. The model requires only 0.03 seconds to generate one second of feedback, underscoring its suitability for real-time applications.

\subsection{Qualitative Evaluation}

In addition to quantitative metrics, we perform qualitative evaluations to assess the perceptual quality and realism of the 3D talking heads generated by DualTalk. These evaluations focus on the accuracy of lip synchronization, smoothness of facial expressions, and the relevance of head and facial movements to the conversational context. We compare the results against several baseline methods, including FaceFormer~\cite{fan2022faceformer}, CodeTalker~\cite{xing2023codetalker}, and EmoTalk~\cite{peng2023emotalk}.

We visualize the output from each method in both speaking and listening modes, as shown in Fig.~\ref{fig:com}. In speaking mode, DualTalk demonstrates larger facial movements and greater expressiveness. Compared to CodeTalker~\cite{xing2023codetalker}, DualTalk achieves notably better visual quality in generating speaking expressions. In listening mode, we visualize a sequence of four frames showing a positive response, where DualTalk effectively combines a smiling expression with a nodding motion. This result underscores DualTalk’s ability to enhance expressiveness in the listener role while providing contextually appropriate responses to the other speaker’s speech and expressions.

To further validate these observations, we conduct a user study in which participants rated the realism and expressiveness of the generated animations. We extract 30 video clips, each lasting over 10 seconds, and invited 30 participants to evaluate them. The questionnaire is designed using the Mean Opinion Score (MOS) rating protocol, asking participants to rate the generated videos from four perspectives: (1) Lip Sync Accuracy, (2) Pose Naturalness, (3) Expression Richness, and (4) Visual Quality. The results are summarized in Tab.~\ref{tab:user}, where DualTalk outperforms previous methods across all evaluations.

\subsection{Ablation Study}

To investigate the contributions of each component in our model, we conduct an ablation study by systematically removing or modifying key modules and inputs. The results, presented in Tab.~\ref{tab:ablation}, highlight the impact of each component on performance, measured by Fréchet Distance (FD), Paired Fréchet Distance (P-FD), Mean Squared Error (MSE), and SI for Diversity (SID) across expression, jaw, and pose.

Removing Speaker-A’s speech and expression leads to significant performance decreases, with FD scores rising to 23.27 and 28.43 for expressions, respectively, emphasizing the importance of incorporating these cues for realistic interactions. Replacing our audio encoder with MFCC results in a decrease in SID from 3.48 to 2.71, demonstrating the effectiveness of our audio encoder in capturing dual-speaker nuances. Excluding the Cross-Modal Temporal Enhancer and Dual-Speaker Interaction Module results in elevated P-FD and MSE scores, highlighting their critical roles in ensuring temporal synchronization and capturing interactive dynamics. Finally, removing the Adaptive Expression Modulation reduces expressiveness, with FD in expressions increasing to 13.28, confirming its value in producing contextually responsive expressions.

\section{Conclusion}

In this paper, we present DualTalk, a unified framework for muti-round dual-speaker 3D talking head generation that seamlessly models both speaker and listener roles. By integrating these roles within a single model, DualTalk enables natural transitions and more realistic interactions in extended conversations. To support this task, we created a large-scale dataset with dual-channel audio and multi-round interactions, providing a benchmark for dual-speaker modeling. Our extensive experiments demonstrated that DualTalk outperforms state-of-the-art methods in both lip synchronization and listener feedback generation, producing more fluid and expressive conversations.

\section*{Acknowledgments}
This research was funded by the National Natural Science Foundation of China under Grant Nos. 62436010 and 62172421, the Beijing Natural Science Foundation under Grant No. 4254100, the Fundamental Research Funds for the Central Universities under Grant No. KG16336301, and the China Postdoctoral Science Foundation under Grant No. 2024M764093. This work was also supported by the Outstanding Innovative Talents Cultivation Funded Programs 2023 of Renmin University of China and the Ant Group Research Intern Program. The project leader was Yanbo Fan.

{
    \small
    \bibliographystyle{ieeenat_fullname}
    \bibliography{main}

\begin{thebibliography}{51}
\providecommand{\natexlab}[1]{#1}
\providecommand{\url}[1]{\texttt{#1}}
\expandafter\ifx\csname urlstyle\endcsname\relax
  \providecommand{\doi}[1]{doi: #1}\else
  \providecommand{\doi}{doi: \begingroup \urlstyle{rm}\Url}\fi

\bibitem[Baevski et~al.(2020)Baevski, Zhou, Mohamed, and Auli]{baevski2020wav2vec}
Alexei Baevski, Yuhao Zhou, Abdelrahman Mohamed, and Michael Auli.
\newblock wav2vec 2.0: A framework for self-supervised learning of speech representations.
\newblock \emph{Advances in neural information processing systems}, 33:\penalty0 12449--12460, 2020.

\bibitem[Bai et~al.(2023)Bai, Fan, Wang, Zhang, Sun, Yuan, and Shan]{bai2023high}
Yunpeng Bai, Yanbo Fan, Xuan Wang, Yong Zhang, Jingxiang Sun, Chun Yuan, and Ying Shan.
\newblock High-fidelity facial avatar reconstruction from monocular video with generative priors.
\newblock In \emph{Proceedings of the IEEE/CVF Conference on Computer Vision and Pattern Recognition}, pages 4541--4551, 2023.

\bibitem[Bredin et~al.(2020)Bredin, Yin, Coria, Gelly, Korshunov, Lavechin, Fustes, Titeux, Bouaziz, and Gill]{bredin2020pyannote}
Herv{\'e} Bredin, Ruiqing Yin, Juan~Manuel Coria, Gregory Gelly, Pavel Korshunov, Marvin Lavechin, Diego Fustes, Hadrien Titeux, Wassim Bouaziz, and Marie-Philippe Gill.
\newblock Pyannote. audio: neural building blocks for speaker diarization.
\newblock In \emph{ICASSP 2020-2020 IEEE International Conference on Acoustics, Speech and Signal Processing (ICASSP)}, pages 7124--7128. IEEE, 2020.

\bibitem[Cao et~al.(2005)Cao, Tien, Faloutsos, and Pighin]{cao2005expressive}
Yong Cao, Wen~C Tien, Petros Faloutsos, and Fr{\'e}d{\'e}ric Pighin.
\newblock Expressive speech-driven facial animation.
\newblock \emph{ACM Transactions on Graphics (TOG)}, 24\penalty0 (4):\penalty0 1283--1302, 2005.

\bibitem[Cassell et~al.(2000)Cassell, Bickmore, Campbell, Vilhjalmsson, Yan, et~al.]{cassell2000human}
Justine Cassell, Tim Bickmore, Lee Campbell, Hannes Vilhjalmsson, Hao Yan, et~al.
\newblock Human conversation as a system framework: Designing embodied conversational agents.
\newblock \emph{Embodied conversational agents}, pages 29--63, 2000.

\bibitem[Chen et~al.()Chen, Zhang, Zhang, Liu, Zhuang, Wan, ZHANG, Li, et~al.]{chencafe}
Hejia Chen, Haoxian Zhang, Shoulong Zhang, Xiaoqiang Liu, Sisi Zhuang, Pengfei Wan, Di ZHANG, Shuai Li, et~al.
\newblock Cafe-talk: Generating 3d talking face animation with multimodal coarse-and fine-grained control.
\newblock In \emph{The Thirteenth International Conference on Learning Representations}.

\bibitem[Cudeiro et~al.(2019)Cudeiro, Bolkart, Laidlaw, Ranjan, and Black]{cudeiro2019capture}
Daniel Cudeiro, Timo Bolkart, Cassidy Laidlaw, Anurag Ranjan, and Michael~J Black.
\newblock Capture, learning, and synthesis of 3d speaking styles.
\newblock In \emph{Proceedings of the IEEE/CVF conference on computer vision and pattern recognition}, pages 10101--10111, 2019.

\bibitem[Diederich et~al.(2022)Diederich, Brendel, Morana, and Kolbe]{diederich2022design}
Stephan Diederich, Alfred~Benedikt Brendel, Stefan Morana, and Lutz Kolbe.
\newblock On the design of and interaction with conversational agents: An organizing and assessing review of human-computer interaction research.
\newblock \emph{Journal of the Association for Information Systems}, 23\penalty0 (1):\penalty0 96--138, 2022.

\bibitem[Fan et~al.(2024)Fan, Li, Lin, Xiao, and Yang]{fan2024unitalker}
Xiangyu Fan, Jiaqi Li, Zhiqian Lin, Weiye Xiao, and Lei Yang.
\newblock Unitalker: Scaling up audio-driven 3d facial animation through a unified model.
\newblock \emph{arXiv preprint arXiv:2408.00762}, 2024.

\bibitem[Fan et~al.(2022)Fan, Lin, Saito, Wang, and Komura]{fan2022faceformer}
Yingruo Fan, Zhaojiang Lin, Jun Saito, Wenping Wang, and Taku Komura.
\newblock Faceformer: Speech-driven 3d facial animation with transformers.
\newblock In \emph{Proceedings of the IEEE/CVF Conference on Computer Vision and Pattern Recognition}, pages 18770--18780, 2022.

\bibitem[Fanelli et~al.(2010)Fanelli, Gall, Romsdorfer, Weise, and Van~Gool]{fanelli20103}
Gabriele Fanelli, Juergen Gall, Harald Romsdorfer, Thibaut Weise, and Luc Van~Gool.
\newblock A 3-d audio-visual corpus of affective communication.
\newblock \emph{IEEE Transactions on Multimedia}, 12\penalty0 (6):\penalty0 591--598, 2010.

\bibitem[Geng et~al.(2023)Geng, Teotia, Tendulkar, Menon, and Vondrick]{geng2023affective}
Scott Geng, Revant Teotia, Purva Tendulkar, Sachit Menon, and Carl Vondrick.
\newblock Affective faces for goal-driven dyadic communication.
\newblock \emph{arXiv preprint arXiv:2301.10939}, 2023.

\bibitem[Gowda et~al.(2023)Gowda, Pandey, and Gowda]{gowda2023pixels}
Shreyank~N Gowda, Dheeraj Pandey, and Shashank~Narayana Gowda.
\newblock From pixels to portraits: A comprehensive survey of talking head generation techniques and applications.
\newblock \emph{arXiv preprint arXiv:2308.16041}, 2023.

\bibitem[Hochreiter(1997)]{hochreiter1997long}
S Hochreiter.
\newblock Long short-term memory.
\newblock \emph{Neural Computation MIT-Press}, 1997.

\bibitem[Huang et~al.(2019)Huang, Fukuda, and Nishida]{huang2019toward}
Hung-Hsuan Huang, Masato Fukuda, and Toyoaki Nishida.
\newblock Toward rnn based micro non-verbal behavior generation for virtual listener agents.
\newblock In \emph{Social Computing and Social Media. Design, Human Behavior and Analytics: 11th International Conference, SCSM 2019, Held as Part of the 21st HCI International Conference, HCII 2019, Orlando, FL, USA, July 26-31, 2019, Proceedings, Part I 21}, pages 53--63. Springer, 2019.

\bibitem[Karras et~al.(2017)Karras, Aila, Laine, Herva, and Lehtinen]{karras2017audio}
Tero Karras, Timo Aila, Samuli Laine, Antti Herva, and Jaakko Lehtinen.
\newblock Audio-driven facial animation by joint end-to-end learning of pose and emotion.
\newblock \emph{ACM Transactions on Graphics (TOG)}, 36\penalty0 (4):\penalty0 1--12, 2017.

\bibitem[Kingma and Ba(2014)]{Kingma2014AdamAM}
Diederik~P. Kingma and Jimmy Ba.
\newblock Adam: A method for stochastic optimization.
\newblock \emph{CoRR}, abs/1412.6980, 2014.

\bibitem[Li et~al.(2024)Li, Yang, Sun, and Hu]{li2024iianet}
Kai Li, Runxuan Yang, Fuchun Sun, and Xiaolin Hu.
\newblock Iianet: An intra-and inter-modality attention network for audio-visual speech separation.
\newblock In \emph{Forty-first International Conference on Machine Learning}, 2024.

\bibitem[Li et~al.(2017)Li, Bolkart, Black, Li, and Romero]{li2017learning}
Tianye Li, Timo Bolkart, Michael~J Black, Hao Li, and Javier Romero.
\newblock Learning a model of facial shape and expression from 4d scans.
\newblock \emph{ACM Trans. Graph.}, 36\penalty0 (6):\penalty0 194--1, 2017.

\bibitem[Li et~al.(2023)Li, Zhang, Wang, Zhao, Wang, Chen, Zhang, Wang, Bo, and Li]{li2023one}
Weichuang Li, Longhao Zhang, Dong Wang, Bin Zhao, Zhigang Wang, Mulin Chen, Bang Zhang, Zhongjian Wang, Liefeng Bo, and Xuelong Li.
\newblock One-shot high-fidelity talking-head synthesis with deformable neural radiance field.
\newblock In \emph{Proceedings of the IEEE/CVF Conference on Computer Vision and Pattern Recognition}, pages 17969--17978, 2023.

\bibitem[Liao et~al.(2023)Liao, Yang, and Shah]{liao2023proactive}
Lizi Liao, Grace~Hui Yang, and Chirag Shah.
\newblock Proactive conversational agents in the post-chatgpt world.
\newblock In \emph{Proceedings of the 46th International ACM SIGIR Conference on Research and Development in Information Retrieval}, pages 3452--3455, 2023.

\bibitem[Liu et~al.(2023)Liu, Wang, Fu, Chai, Yu, Dai, and Han]{liu2023mfr}
Jin Liu, Xi Wang, Xiaomeng Fu, Yesheng Chai, Cai Yu, Jiao Dai, and Jizhong Han.
\newblock Mfr-net: Multi-faceted responsive listening head generation via denoising diffusion model.
\newblock In \emph{Proceedings of the 31st ACM International Conference on Multimedia}, pages 6734--6743, 2023.

\bibitem[Liu et~al.(2024)Liu, Guo, Zhen, Li, Ao, and Yan]{liu2024customlistener}
Xi Liu, Ying Guo, Cheng Zhen, Tong Li, Yingying Ao, and Pengfei Yan.
\newblock Customlistener: Text-guided responsive interaction for user-friendly listening head generation.
\newblock In \emph{Proceedings of the IEEE/CVF Conference on Computer Vision and Pattern Recognition}, pages 2415--2424, 2024.

\bibitem[Lugaresi et~al.(2019)Lugaresi, Tang, Nash, McClanahan, Uboweja, Hays, Zhang, Chang, Yong, Lee, et~al.]{lugaresi2019mediapipe}
Camillo Lugaresi, Jiuqiang Tang, Hadon Nash, Chris McClanahan, Esha Uboweja, Michael Hays, Fan Zhang, Chuo-Ling Chang, Ming~Guang Yong, Juhyun Lee, et~al.
\newblock Mediapipe: A framework for building perception pipelines.
\newblock \emph{arXiv preprint arXiv:1906.08172}, 2019.

\bibitem[Luo et~al.(2023)Luo, Song, Xie, Spitale, Shen, and Gunes]{luo2023reactface}
Cheng Luo, Siyang Song, Weicheng Xie, Micol Spitale, Linlin Shen, and Hatice Gunes.
\newblock Reactface: Multiple appropriate facial reaction generation in dyadic interactions.
\newblock \emph{arXiv preprint arXiv:2305.15748}, 2023.

\bibitem[Ma et~al.(2024)Ma, Zhu, Qi, Qian, Zhang, and Lei]{ma2024diffspeaker}
Zhiyuan Ma, Xiangyu Zhu, Guojun Qi, Chen Qian, Zhaoxiang Zhang, and Zhen Lei.
\newblock Diffspeaker: Speech-driven 3d facial animation with diffusion transformer.
\newblock \emph{arXiv preprint arXiv:2402.05712}, 2024.

\bibitem[Ng et~al.(2022)Ng, Joo, Hu, Li, Darrell, Kanazawa, and Ginosar]{ng2022learning}
Evonne Ng, Hanbyul Joo, Liwen Hu, Hao Li, Trevor Darrell, Angjoo Kanazawa, and Shiry Ginosar.
\newblock Learning to listen: Modeling non-deterministic dyadic facial motion.
\newblock In \emph{Proceedings of the IEEE/CVF Conference on Computer Vision and Pattern Recognition}, pages 20395--20405, 2022.

\bibitem[Ng et~al.(2023)Ng, Subramanian, Klein, Kanazawa, Darrell, and Ginosar]{ng2023can}
Evonne Ng, Sanjay Subramanian, Dan Klein, Angjoo Kanazawa, Trevor Darrell, and Shiry Ginosar.
\newblock Can language models learn to listen?
\newblock In \emph{Proceedings of the IEEE/CVF International Conference on Computer Vision}, pages 10083--10093, 2023.

\bibitem[Ng et~al.(2024)Ng, Romero, Bagautdinov, Bai, Darrell, Kanazawa, and Richard]{ng2024audio}
Evonne Ng, Javier Romero, Timur Bagautdinov, Shaojie Bai, Trevor Darrell, Angjoo Kanazawa, and Alexander Richard.
\newblock From audio to photoreal embodiment: Synthesizing humans in conversations.
\newblock In \emph{Proceedings of the IEEE/CVF Conference on Computer Vision and Pattern Recognition}, pages 1001--1010, 2024.

\bibitem[Niswar et~al.(2009)Niswar, Ong, Nguyen, and Huang]{niswar2009real}
Arthur Niswar, Ee~Ping Ong, Hong~Thai Nguyen, and Zhiyong Huang.
\newblock Real-time 3d talking head from a synthetic viseme dataset.
\newblock In \emph{Proceedings of the 8th International Conference on Virtual Reality Continuum and its Applications in Industry}, pages 29--33, 2009.

\bibitem[Nocentini et~al.(2024)Nocentini, Besnier, Ferrari, Arguillere, Berretti, and Daoudi]{nocentini2024scantalk}
Federico Nocentini, Thomas Besnier, Claudio Ferrari, Sylvain Arguillere, Stefano Berretti, and Mohamed Daoudi.
\newblock Scantalk: 3d talking heads from unregistered scans.
\newblock \emph{arXiv preprint arXiv:2403.10942}, 2024.

\bibitem[Pang et~al.(2023)Pang, Zhang, Quan, Fan, Cun, Shan, and Yan]{pang2023dpe}
Youxin Pang, Yong Zhang, Weize Quan, Yanbo Fan, Xiaodong Cun, Ying Shan, and Dong-ming Yan.
\newblock Dpe: Disentanglement of pose and expression for general video portrait editing.
\newblock In \emph{Proceedings of the IEEE/CVF Conference on Computer Vision and Pattern Recognition}, pages 427--436, 2023.

\bibitem[Peng et~al.(2023{\natexlab{a}})Peng, Luo, Shi, Xu, Zhu, Liu, He, and Fan]{peng2023selftalk}
Ziqiao Peng, Yihao Luo, Yue Shi, Hao Xu, Xiangyu Zhu, Hongyan Liu, Jun He, and Zhaoxin Fan.
\newblock Selftalk: A self-supervised commutative training diagram to comprehend 3d talking faces.
\newblock In \emph{Proceedings of the 31st ACM International Conference on Multimedia}, pages 5292--5301, 2023{\natexlab{a}}.

\bibitem[Peng et~al.(2023{\natexlab{b}})Peng, Wu, Song, Xu, Zhu, He, Liu, and Fan]{peng2023emotalk}
Ziqiao Peng, Haoyu Wu, Zhenbo Song, Hao Xu, Xiangyu Zhu, Jun He, Hongyan Liu, and Zhaoxin Fan.
\newblock Emotalk: Speech-driven emotional disentanglement for 3d face animation.
\newblock In \emph{Proceedings of the IEEE/CVF International Conference on Computer Vision}, pages 20687--20697, 2023{\natexlab{b}}.

\bibitem[Peng et~al.(2024)Peng, Hu, Shi, Zhu, Zhang, Zhao, He, Liu, and Fan]{peng2024synctalk}
Ziqiao Peng, Wentao Hu, Yue Shi, Xiangyu Zhu, Xiaomei Zhang, Hao Zhao, Jun He, Hongyan Liu, and Zhaoxin Fan.
\newblock Synctalk: The devil is in the synchronization for talking head synthesis.
\newblock In \emph{Proceedings of the IEEE/CVF Conference on Computer Vision and Pattern Recognition}, pages 666--676, 2024.

\bibitem[Richard et~al.(2021)Richard, Zollh{\"o}fer, Wen, De~la Torre, and Sheikh]{richard2021meshtalk}
Alexander Richard, Michael Zollh{\"o}fer, Yandong Wen, Fernando De~la Torre, and Yaser Sheikh.
\newblock Meshtalk: 3d face animation from speech using cross-modality disentanglement.
\newblock In \emph{Proceedings of the IEEE/CVF International Conference on Computer Vision}, pages 1173--1182, 2021.

\bibitem[Song et~al.(2023{\natexlab{a}})Song, Yin, Jin, Dong, and Xu]{song2023emotional}
Luchuan Song, Guojun Yin, Zhenchao Jin, Xiaoyi Dong, and Chenliang Xu.
\newblock Emotional listener portrait: Neural listener head generation with emotion.
\newblock In \emph{Proceedings of the IEEE/CVF International Conference on Computer Vision}, pages 20839--20849, 2023{\natexlab{a}}.

\bibitem[Song et~al.(2023{\natexlab{b}})Song, Spitale, Luo, Barquero, Palmero, Escalera, Valstar, Baur, Ringeval, Andr{\'e}, et~al.]{song2023react2023}
Siyang Song, Micol Spitale, Cheng Luo, Germ{\'a}n Barquero, Cristina Palmero, Sergio Escalera, Michel Valstar, Tobias Baur, Fabien Ringeval, Elisabeth Andr{\'e}, et~al.
\newblock React2023: The first multiple appropriate facial reaction generation challenge.
\newblock In \emph{Proceedings of the 31st ACM International Conference on Multimedia}, pages 9620--9624, 2023{\natexlab{b}}.

\bibitem[Sou{\v{c}}ek and Loko{\v{c}}(2020)]{souvcek2020transnet}
Tom{\'a}{\v{s}} Sou{\v{c}}ek and Jakub Loko{\v{c}}.
\newblock Transnet v2: An effective deep network architecture for fast shot transition detection.
\newblock \emph{arXiv preprint arXiv:2008.04838}, 2020.

\bibitem[Sung-Bin et~al.(2024)Sung-Bin, Hyun, Hong, Nam, Ju, and Oh]{sung2024laughtalk}
Kim Sung-Bin, Lee Hyun, Da~Hye Hong, Suekyeong Nam, Janghoon Ju, and Tae-Hyun Oh.
\newblock Laughtalk: Expressive 3d talking head generation with laughter.
\newblock In \emph{Proceedings of the IEEE/CVF Winter Conference on Applications of Computer Vision}, pages 6404--6413, 2024.

\bibitem[Thambiraja et~al.(2023)Thambiraja, Habibie, Aliakbarian, Cosker, Theobalt, and Thies]{thambiraja2023imitator}
Balamurugan Thambiraja, Ikhsanul Habibie, Sadegh Aliakbarian, Darren Cosker, Christian Theobalt, and Justus Thies.
\newblock Imitator: Personalized speech-driven 3d facial animation.
\newblock In \emph{Proceedings of the IEEE/CVF International Conference on Computer Vision}, pages 20621--20631, 2023.

\bibitem[Tran et~al.(2024)Tran, Chang, Siniukov, and Soleymani]{tran2024dyadic}
Minh Tran, Di Chang, Maksim Siniukov, and Mohammad Soleymani.
\newblock Dyadic interaction modeling for social behavior generation.
\newblock \emph{arXiv preprint arXiv:2403.09069}, 2024.

\bibitem[Vaswani(2017)]{vaswani2017attention}
A Vaswani.
\newblock Attention is all you need.
\newblock \emph{Advances in Neural Information Processing Systems}, 2017.

\bibitem[Wu et~al.(2024)Wu, Peng, Zhou, Cheng, He, Liu, and Fan]{wu2024vgg}
Haoyu Wu, Ziqiao Peng, Xukun Zhou, Yunfei Cheng, Jun He, Hongyan Liu, and Zhaoxin Fan.
\newblock Vgg-tex: A vivid geometry-guided facial texture estimation model for high fidelity monocular 3d face reconstruction.
\newblock \emph{arXiv preprint arXiv:2409.09740}, 2024.

\bibitem[Wu et~al.(2023)Wu, Bansal, Zhang, Wu, Zhang, Zhu, Li, Jiang, Zhang, and Wang]{wu2023autogen}
Qingyun Wu, Gagan Bansal, Jieyu Zhang, Yiran Wu, Shaokun Zhang, Erkang Zhu, Beibin Li, Li Jiang, Xiaoyun Zhang, and Chi Wang.
\newblock Autogen: Enabling next-gen llm applications via multi-agent conversation framework.
\newblock \emph{arXiv preprint arXiv:2308.08155}, 2023.

\bibitem[Xing et~al.(2023)Xing, Xia, Zhang, Cun, Wang, and Wong]{xing2023codetalker}
Jinbo Xing, Menghan Xia, Yuechen Zhang, Xiaodong Cun, Jue Wang, and Tien-Tsin Wong.
\newblock Codetalker: Speech-driven 3d facial animation with discrete motion prior.
\newblock In \emph{Proceedings of the IEEE/CVF Conference on Computer Vision and Pattern Recognition}, pages 12780--12790, 2023.

\bibitem[Yu et~al.(2023)Yu, Fan, Zhang, Wang, Yin, Bai, Cao, Shan, Wu, Sun, et~al.]{yu2023nofa}
Wangbo Yu, Yanbo Fan, Yong Zhang, Xuan Wang, Fei Yin, Yunpeng Bai, Yan-Pei Cao, Ying Shan, Yang Wu, Zhongqian Sun, et~al.
\newblock Nofa: Nerf-based one-shot facial avatar reconstruction.
\newblock In \emph{ACM SIGGRAPH 2023 conference proceedings}, pages 1--12, 2023.

\bibitem[Zhen et~al.(2023)Zhen, Song, He, Cao, Shi, and Luo]{zhen2023human}
Rui Zhen, Wenchao Song, Qiang He, Juan Cao, Lei Shi, and Jia Luo.
\newblock Human-computer interaction system: A survey of talking-head generation.
\newblock \emph{Electronics}, 12\penalty0 (1):\penalty0 218, 2023.

\bibitem[Zhou et~al.(2022)Zhou, Bai, Zhang, Yao, Zhao, and Mei]{zhou2022responsive}
Mohan Zhou, Yalong Bai, Wei Zhang, Ting Yao, Tiejun Zhao, and Tao Mei.
\newblock Responsive listening head generation: a benchmark dataset and baseline.
\newblock In \emph{European Conference on Computer Vision}, pages 124--142. Springer, 2022.

\bibitem[Zhou et~al.(2024)Zhou, Li, Peng, Wu, He, Qin, Fan, and Liu]{zhou2024meta}
Xukun Zhou, Fengxin Li, Ziqiao Peng, Kejian Wu, Jun He, Biao Qin, Zhaoxin Fan, and Hongyan Liu.
\newblock Meta-learning empowered meta-face: Personalized speaking style adaptation for audio-driven 3d talking face animation.
\newblock \emph{arXiv preprint arXiv:2408.09357}, 2024.

\bibitem[Zhou et~al.(2018)Zhou, Xu, Landreth, Kalogerakis, Maji, and Singh]{zhou2018visemenet}
Yang Zhou, Zhan Xu, Chris Landreth, Evangelos Kalogerakis, Subhransu Maji, and Karan Singh.
\newblock Visemenet: Audio-driven animator-centric speech animation.
\newblock \emph{ACM Transactions on Graphics (TOG)}, 37\penalty0 (4):\penalty0 1--10, 2018.

\end{thebibliography}
}

\newpage
\setcounter{section}{0}
\maketitlesupplementary
In this supplementary material, we provide additional details on DualTalk. Section 1 covers the implementation details, including network architecture and loss functions. Section 2 describes the dataset collection and processing methods. Section 3 outlines the evaluation metrics used to assess performance. Section 4 discusses ethical considerations, and Section 5 addresses limitations and future work.

\section{Implementation Details}
\subsection{Network Architecture}
In this section, we provide comprehensive implementation details of our DualTalk framework. The framework consists of four main components: Dual-Speaker Joint Encoder, Cross-Modal Temporal Enhancer, Dual-Speaker Interaction Module, and Expressive Synthesis Module.

The Dual-Speaker Joint Encoder processes both audio and visual inputs through parallel branches. For audio processing, we utilize a pre-trained Wav2Vec 2.0~\cite{baevski2020wav2vec} model to encode the raw audio waveforms (sampled at 16kHz) into high-dimensional feature representations. The audio encoder consists of 12 transformer~\cite{vaswani2017attention} layers with a hidden dimension of 1024, followed by a linear projection layer that maps the features to a 256-dimensional space. This projection is essential for aligning the audio features with the visual representation space. The visual branch processes blendshape coefficients through a two-layer MLP with ReLU activations, where the first layer maps the 56 blendshape parameters to 128 dimensions, and the second layer further projects these features to match the 256-dimensional audio features.

The Cross-Modal Temporal Enhancer is designed to ensure temporal coherence and modal alignment. At its core is a multimodal cross-attention mechanism with 4 attention heads. This mechanism allows the model to establish connections between audio and visual features across different temporal positions. Following the cross-attention layer, we employ a bidirectional LSTM~\cite{hochreiter1997long} with 512 hidden units and 2 layers to capture long-term dependencies in both forward and backward directions. The LSTM incorporates a dropout of 0.1 between layers to prevent overfitting.

For the Dual-Speaker Interaction Module, we implement a transformer-based architecture consisting of an encoder and decoder, each with 3 layers. The encoder employs 4-head self-attention mechanisms with a hidden dimension of 256 and a feed-forward network dimension of 512. The Modal Alignment Attention layer, inspired by FaceFormer~\cite{fan2022faceformer}, uses a custom attention mask to ensure causal relationships in the temporal domain. The decoder follows a similar structure but includes additional cross-attention layers to integrate information from both speakers.

The Expressive Synthesis Module utilizes an adaptive expression modulation mechanism implemented as a two-layer MLP. The first layer expands the 256-dimensional features to 512 dimensions, followed by layer normalization and ReLU activation. The second layer then projects back to 256 dimensions before the final blendshape prediction layer, which outputs 56 blendshape parameters normalized through a sigmoid activation.

\subsection{Loss Functions}
Our training objective incorporates multiple loss terms to ensure both accurate blendshape prediction and smooth temporal dynamics. The total loss function consists of two primary components: a direct blendshape reconstruction loss and a velocity loss that enforces temporal consistency.

The blendshape reconstruction loss ($\mathcal{L}_{bs}$) is computed as the Mean Squared Error (MSE) between the predicted head motion blendshape parameters ($\hat{M}$) and the ground truth blendshapes ($M$):

\begin{equation}
\mathcal{L}_{bs} = \text{MSE}(\hat{M}, M) = \frac{1}{N}\sum_{i=1}^{N}(\hat{M}_i - M_i)^2
\end{equation}

To ensure smooth and natural facial movements, we introduce a velocity loss term that penalizes sudden changes in blendshape parameters between consecutive frames. The velocity is computed as the first-order temporal difference of blendshape parameters. Specifically, for both predicted and ground truth sequences, we calculate the frame-to-frame differences:

\begin{equation}
V_{gt} = M_{t+1} - M_t
\end{equation}

\begin{equation}
\hat{V} = \hat{M}_{t+1} - \hat{M}_t
\end{equation}

where $t$ represents the frame index. The velocity loss ($\mathcal{L}_{vel}$) is then computed as the MSE between the predicted and ground truth velocities:

\begin{equation}
\mathcal{L}_{vel} = \text{MSE}(\hat{V}, V_{gt}) = \frac{1}{N-1}\sum_{t=1}^{N-1}(\hat{V}_t - V_{gt,t})^2
\end{equation}

The final loss is the mean of these two components:

\begin{equation}
\mathcal{L}_{total} = \mathcal{L}_{bs} + \mathcal{L}_{vel}
\end{equation}

This combined loss function effectively balances between accurate facial expression reproduction and temporal smoothness. The blendshape reconstruction loss ensures that the predicted facial expressions match the ground truth at each frame, while the velocity loss prevents unrealistic, jittery movements by encouraging smooth transitions between consecutive frames. During training, we use equally weighting these two terms (with an implicit weight of 1.0 for each).

\subsection{Training Details}
During training, we optimize our model using the Adam~\cite{Kingma2014AdamAM} optimizer with an initial learning rate of 1e-4. We train the model for 200 epochs using a batch size of 32 on a NVIDIA A6000 GPU with 48GB memory each. The complete training process takes approximately 48 hours to converge.

\section{Dataset Details}

Our dataset collection and processing pipeline is designed to create a comprehensive and high-quality dataset for dual-speaker interaction modeling. Here, we provide detailed information about our data collection, processing procedures, and dataset statistics.

The raw data is collected from YouTube interviews, with a wide variety of natural face-to-face interactions. We specifically focus on videos featuring clear facial visibility of both speakers, high-quality audio, and natural conversational dynamics. All videos are in 1920×1080 resolution recorded at 25 frames per second, with audio sampled at 16kHz. The collected videos span different languages, speaking styles, and environmental conditions to ensure robustness and generalization of our model.

The resulting dataset comprises 50 hours of processed conversation data, featuring 1,052 unique identities across 5,858 video clips. Each clip contains an average of 2.5 conversation rounds, where speakers naturally alternate between speaking and listening roles. The dataset is carefully split into training (4,935 clips), testing (539 clips), and out-of-distribution (OOD) validation sets (384 clips). The OOD set specifically includes speakers and conversation scenarios not present in the training data to evaluate generalization capability.

To construct this dataset, we sourced two-person conversational videos from YouTube and RealTalk~\cite{geng2023affective} raw videos. 
Videos are segmented using TransNet V2~\cite{souvcek2020transnet} for shot transition detection, retaining only segments longer than 5 seconds to capture meaningful interactions. Visual-guided speech separation is performed with IIANet~\cite{li2024iianet}, producing isolated audio streams for each speaker—a critical feature for accurate lip synchronization and expression modeling.

To ensure speaker-specific frame isolation, we use MediaPipe~\cite{lugaresi2019mediapipe} for face detection and tracking. High-resolution 3D facial meshes are extracted using Spectre, and samples with abnormal coefficients are filtered out. For speaker separation, Pyannote~\cite{bredin2020pyannote} is employed, allowing the identification of multi-round conversations and distinct speaker turns to facilitate the extraction of back-and-forth dialogues. To ensure annotation stability, a minimum speech duration of 2 seconds is set.

\section{Evaluation Metrics}

In this section, we provide detailed descriptions of the evaluation metrics used to assess the performance of our DualTalk framework. These metrics are carefully selected to comprehensively evaluate different aspects of the generated conversational animations, including motion realism, temporal synchronization, and interaction dynamics.

\textbf{Fréchet Distance (FD):} The FD serves as our primary metric for evaluating motion realism. It computes the distributional distance between generated and ground-truth motions in the feature space. Specifically, we extract deep features from both the predicted and actual motion sequences using a pre-trained motion encoder, modeling them as multivariate Gaussian distributions. The FD effectively captures the statistical similarity between the generated and real motion distributions, where a lower score indicates better motion realism.

\textbf{Paired Fréchet Distance (P-FD):} To evaluate the quality of dual-speaker interactions, we introduce the P-FD metric, which extends the traditional FD by considering the joint distribution of dual-speaker pairs. By concatenating the generated Speaker-B's motions with the corresponding Speaker-A's motions along the feature dimension, we compute the FD between these paired representations and their ground-truth counterparts. This approach captures the synchronization and coherence between the two speakers’ movements, providing insights into the quality of interactive dynamics.

\textbf{Mean Squared Error (MSE):} For direct motion accuracy assessment, we employ the MSE between generated and ground-truth motions. This metric is computed across all blendshape parameters and temporal dimensions, providing a straightforward measure of prediction accuracy. The MSE helps us understand how closely the generated animations match the ground truth at a frame-by-frame level.

\textbf{SI for Diversity (SID):} To evaluate the diversity of generated animations, we use the SID metric. This approach applies k-means clustering (k=40) to the motion sequences in the feature space and quantifies diversity by calculating the entropy of the cluster assignment histogram. A higher SID value indicates more diverse and varied motion patterns in the generated animations, which is crucial for producing natural and non-repetitive conversational behaviors.

\textbf{Residual Pearson Correlation Coefficient (rPCC):} To assess the temporal correlation between Speaker-A and Speaker-B movements, we introduce the rPCC metric. It computes the frame-wise Pearson correlation between Speaker-A and Speaker-B motions and then measures the L1 distance between the correlation patterns of generated and ground-truth sequences. The rPCC is particularly useful for evaluating how well the model captures the subtle interactive dynamics between Speaker-A and Speaker-B in conversation.

These metrics collectively provide a comprehensive evaluation framework for assessing the quality, realism, and interactive dynamics of our dual-speaker animation system. Each metric focuses on a specific aspect of the generated animations, enabling detailed analysis of the model's performance across different dimensions. Through this multi-faceted evaluation approach, we can thoroughly validate the effectiveness of our proposed method in generating realistic and interactive conversational animations.

\section{Ethics Considerations}
The development of DualTalk raises important ethical considerations, particularly regarding privacy, misuse, and potential societal impacts. The DualTalk dataset includes extensive conversational data, and while publicly available sources were used, ensuring compliance with data privacy laws and ethical guidelines remains a priority. Steps have been taken to anonymize and process data responsibly, but future work will aim to establish more robust safeguards to prevent inadvertent exposure of personal information.

Another key concern is the potential misuse of DualTalk for deceptive purposes, such as creating realistic yet fabricated conversations or impersonating individuals. To mitigate this, strict usage policies and watermarking techniques can be implemented to differentiate generated content from real-world interactions. Open-sourcing the technology will be accompanied by clear guidelines to discourage unethical applications.

\section{Limitations and Future Works}

The limitations of DualTalk primarily lie in its current focus on dyadic interactions and the lack of precise emotional controllability in generated animations. While DualTalk excels in creating synchronized and natural two-speaker conversations, it cannot yet handle multi-party interactions, which are common in real-world applications. Additionally, while the Expressive Synthesis Module generates nuanced facial expressions, the model lacks the ability to precisely control the emotional tone of its outputs, limiting its adaptability to specific scenarios or user preferences.

Future work will focus on extending DualTalk to multi-party interactions, enabling the model to handle dynamic role transitions and conversational flows in group settings. Additionally, efforts will be directed toward generating controllable emotions, allowing the system to adapt its responses to specific emotional tones or user preferences, further enhancing the naturalness and versatility of 3D talking head animations.

\end{document}